\documentclass[10pt,twocolumn,letterpaper]{article}

\usepackage{cvpr}
\usepackage{times}
\usepackage{epsfig}
\usepackage{graphicx}
\usepackage{amsmath}
\usepackage{amssymb}

\usepackage{array}
\usepackage{color}
\usepackage{multirow}
\usepackage{hhline}
\usepackage{subfigure}
\usepackage{algorithm}
\usepackage{algorithmic}
\usepackage{amsmath}
\usepackage{amssymb}
\usepackage{amsfonts}
\usepackage{amsthm}
\usepackage{booktabs}
\usepackage{subfigure}
\usepackage{listings}
\usepackage{soul}

\makeatletter
\newcommand*{\rom}[1]{\expandafter\@slowromancap\romannumeral #1@}
\makeatother

\usepackage[pagebackref=true,breaklinks=true,letterpaper=true,colorlinks,bookmarks=false]{hyperref}

\cvprfinalcopy 

\def\cvprPaperID{1821} 

\begin{document}
\graphicspath{{figures/}}

\title{Real-Time Referring Expression Comprehension by \\ Single-Stage Grounding Network}

\author{Xinpeng Chen$^{1}$\thanks{Work done while Xinpeng Chen and Jingyuan Chen were Research Interns with Tencent AI Lab.} \qquad Lin Ma$^1$\thanks{Corresponding author.} \qquad Jingyuan Chen$^{2}$$^\ast$ \qquad Zequn Jie$^{1}$ \qquad Wei Liu$^1$ \qquad Jiebo Luo$^{3}$ \\
$^1$Tencent AI Lab \qquad  $^2$National University of Singapore \qquad $^{3}$University of Rochester\\
{\tt\small \{jschenxinpeng, forest.linma, jingyuanchen91, zequn.nus\}@gmail.com} \\
{\tt\small wl2223@columbia.edu} \qquad
{\tt\small jluo@cs.rochester.edu}}

\maketitle
\begin{abstract}
In this paper, we propose a novel end-to-end model, namely Single-Stage Grounding network~(SSG), to localize the referent given a referring expression within an image. Different from previous multi-stage models which rely on object proposals or detected regions, our proposed model aims to comprehend a referring expression through one single stage without resorting to region proposals as well as the subsequent region-wise feature extraction. Specifically, a multimodal interactor is proposed to summarize the local region features regarding the referring expression attentively. Subsequently, a grounder is proposed to localize the referring expression within the given image directly. For further improving the localization accuracy, a guided attention mechanism is proposed to enforce the grounder to focus on the central region of the referent. Moreover, by exploiting and predicting visual attribute information, the grounder can further distinguish the referent objects within an image and thereby improve the model performance. Experiments on RefCOCO, RefCOCO+, and RefCOCOg datasets demonstrate that our proposed SSG without relying on any region proposals can achieve comparable performance with other advanced models. Furthermore, our SSG outperforms the previous models and achieves the state-of-art performance on the ReferItGame dataset. More importantly, our SSG is time efficient and can ground a referring expression in a $416 \times 416$ image from the RefCOCO dataset in 25ms~(40 referents per second) on average with a Nvidia Tesla P40, accomplishing more than 9$\times$ speedups over the existing multi-stage models.
\end{abstract}
\vspace{-5pt}





\begin{figure}[!t]
  \centering
  \includegraphics[width=\linewidth]{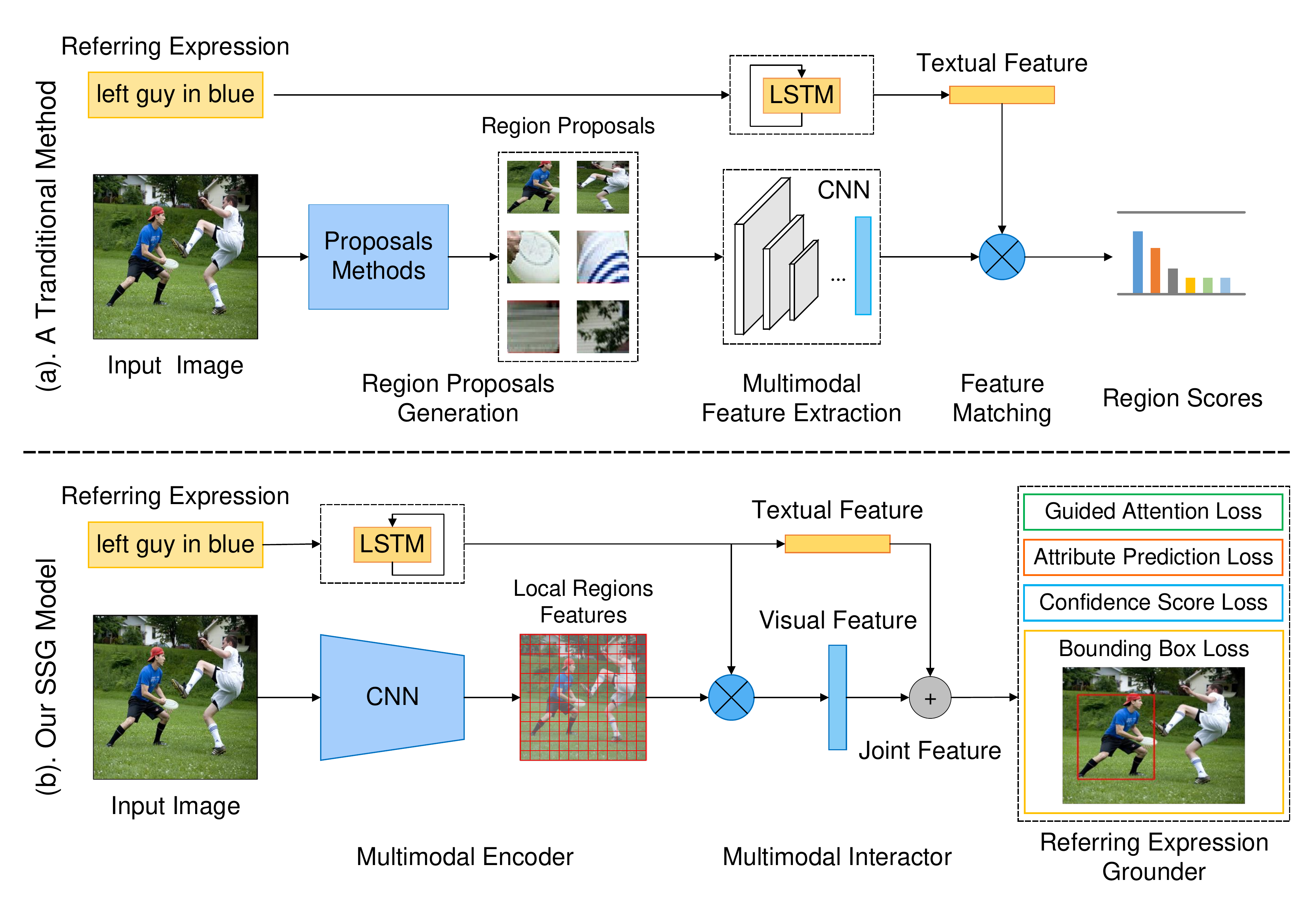}
  \caption{A comparison between our SSG model and a traditional multi-stage method. By completely discarding the region proposal generation stage and directly predicting the bounding box for the referring expression, our SSG model runs faster by design.}
  \vspace{-10pt}
  \label{fig_intro}
\end{figure}

\section{Introduction}
\label{sec:intro}
The referring expression comprehension~\cite{yu2018mattnet,yu2016modeling,yu_cvpr2017_joint,zhang2017grounding}, also known as referring expression grounding, is a fundamental research problem which has received increasing attention from both computer vision and natural language processing research communities. Given an image as well as a referring expression, which describes a specific referent within the image, the referring expression comprehension aims to localize the referent corresponding to the semantic meaning of the referring expression. This is a general-purpose yet challenging vision plus language task, since it requires not only localization of the referent, but also high-level semantic comprehension of the referring and relationships (\textit{e.g.} ``left" in Fig.~\ref{fig_intro}) that help distinguish the correct referent from the other unrelated ones in the same image.

Previous referring expression comprehension models can be regarded as multi-stage methods which comprise three stages~\cite{SCRC,mao_MMI,nagaraja16refexp,GroundR,yu2018mattnet,yu2016modeling,yu_cvpr2017_joint,zhang2017grounding}, as illustrated in Fig.~\ref{fig_intro}~(a). First, the conventional object proposal generation methods, such as EdgeBox~\cite{edgeboxes}, Selective Search~\cite{selectivesearch}, or off-the-shelf object detectors such as Faster R-CNN~\cite{faster-rcnn}, SSD~\cite{ssd}, and mask R-CNN~\cite{he2017maskrcnn}, are utilized to extract a set of regions as the candidates for matching the referring expression. Second, convolutional neural networks (CNNs)~\cite{vgg,inception-v4} and recurrent neural networks (RNNs)~\cite{Cho_2014,Schmidhuber_1997} are used to encode the image regions and the referring expression, respectively. Finally, a ranking model is designed to select the region with the highest matching score as the referent. These multi-stage models have achieved remarkable performance over related datasets on the referring expression comprehension task~\cite{yu2018mattnet,yu_cvpr2017_joint,zhang2017grounding}.

However, these multi-stage models are very computationally expensive, with high time cost taken in each stage, especially region proposal generation and region-wise feature extraction, as illustrated in Table~\ref{Table_speed}. As such, these models are not applicable to the practical scenarios with real-time requirements. Therefore, this new challenge motivates and inspires us to design a grounding model which can localize the referent within an image both effectively and efficiently. To this end, in this paper, we propose a Single-Stage Grounding network~(SSG) to achieve the real-time grounding results as well as the favorable performance without resorting to region proposals. More specifically, as shown in Fig.~\ref{fig_intro}~(b), our SSG model consists of three components, namely multimodal encoder, multimodal interactor, and referring expression grounder. The multimodal encoder~(Sec.~\ref{sec:encoder}) is leveraged to encode the given image and the referring expression, respectively. The multimodal interactor~(Sec.~\ref{sec:interactor}) aims to attentively summarize the image local representations conditioned on the textual representation. Finally, based on the joint representation, the referring expression grounder~(Sec.~\ref{sec:grounder}) is responsible for directly predicting the coordinates of the bounding box corresponding to the referring expression. In addition to the bounding box regression loss, additional three auxiliary losses are introduced to further improve the performance of SSG. They are the confidence score loss (Sec.~\ref{sec:localization}) reflecting how accurate the bounding box is, the attention weight loss (Sec.~\ref{sec:guided-attention}) enforcing the grounder to focus on the useful region by using the central point of the ground-truth bounding box as the target, and the attribute prediction loss (Sec.~\ref{sec:attribute-prediction}) benefiting to distinguish the referring objects in the same image. As such, our [proposed SSG performs in one single stage for tackling the referring expression comprehension task, thus leading to the comparable model performance as well as more than 9$\times$ speedups over the existing multi-stage models.

In summary, the main contributions of our work are as follows:
\begin{itemize}
    \item We propose a novel end-to-end model, namely Single-Stage Grounding network~(SSG) for addressing the referring expression comprehension task, which directly predicts the coordinates of the bounding box within the given image corresponding to the referring expression without relying on any region proposals. 
    \item We propose a guided attention mechanism with the object center-bias to encourage our SSG to focus on the central region of a referent. Moreover, our proposed SSG can further distinguish referent objects, by exploiting and predicting the visual attribute information.  
    \item Our SSG can carry out the referring expression comprehension task both effectively and efficiently. Specifically, our SSG achieves comparable results with the state-of-the-art models, while taking more than 9$\times$ faster under the same hardware environment.
\end{itemize}






\section{Related Work}

\subsection{Referring Expression Comprehension}
The referring expression comprehension task is to localize a referent within the given image, which semantically corresponds to the given referring expression. This task involves comprehending and modeling the different spatial contexts, such as spatial configurations~\cite{mao_MMI,yu2016modeling}, attributes~\cite{liu_iccv17_attr,yu2018mattnet}, and the relationships between regions~\cite{nagaraja16refexp,yu2016modeling}. In previous work, this task is generally formulated as a ranking problem over a set of region proposals from the given image. The region proposals are extracted from the proposal generation methods such as EdgeBoxes~\cite{edgeboxes}, or advanced object detection methods such as SSD~\cite{ssd}, Faster RCNN~\cite{faster-rcnn}, and Mask R-CNN~\cite{he2017maskrcnn}. Earlier models~\cite{mao_MMI,yu2016modeling} scored region proposals according to  visual and spatial feature representations. However, these methods fail to incorporate the interactions between objects because the scoring process of each region is isolated. Nagaraja et al.~\cite{nagaraja16refexp} improved the performance with the help of modeling the relationships between region proposals. Yu et al.~\cite{yu_cvpr2017_joint} proposed a joint framework that integrates referring expression comprehension and generation tasks together. The visual features from the region proposals and the semantic information from the referring expressions are embedded into a common space. Zhang et al.~\cite{zhang2017grounding} developed a variational Bayesian framework to exploit the reciprocity between the referent and context. In spite of these models and their variants have achieved remarkable performance improvements on the referring comprehension task~\cite{yu2018mattnet}, these multi-stage methods could be computationally expensive for practical applications. 

\subsection{Object Detection}
Our proposed SSG also benefits from the state-of-art object detectors, especially YOLO~\cite{yolo}, YOLO-v2~\cite{yolo2}, and YOLO-v3~\cite{yolo3}. YOLO~\cite{yolo} divides an input image into $7 \times 7$ grid cells and directly predicts both the confidence values for multiple categories and coordinates of the bounding boxes. Similar to YOLO, YOLO-v2~\cite{yolo2} also divides an input image into a set of grid cells. However, it places 5 anchor boxes at each grid cell and predicts corrections of the anchor boxes. Furthermore, YOLO-v3 takes a deeper network with 53 convolutional layers as the backbone which is more powerful. In order to localize small objects, YOLO-v3~\cite{yolo3} also introduces the additional pass-through layer to obtain more fine-grained features.

\begin{figure*}[!t]
    \centering
    \includegraphics[width=\linewidth]{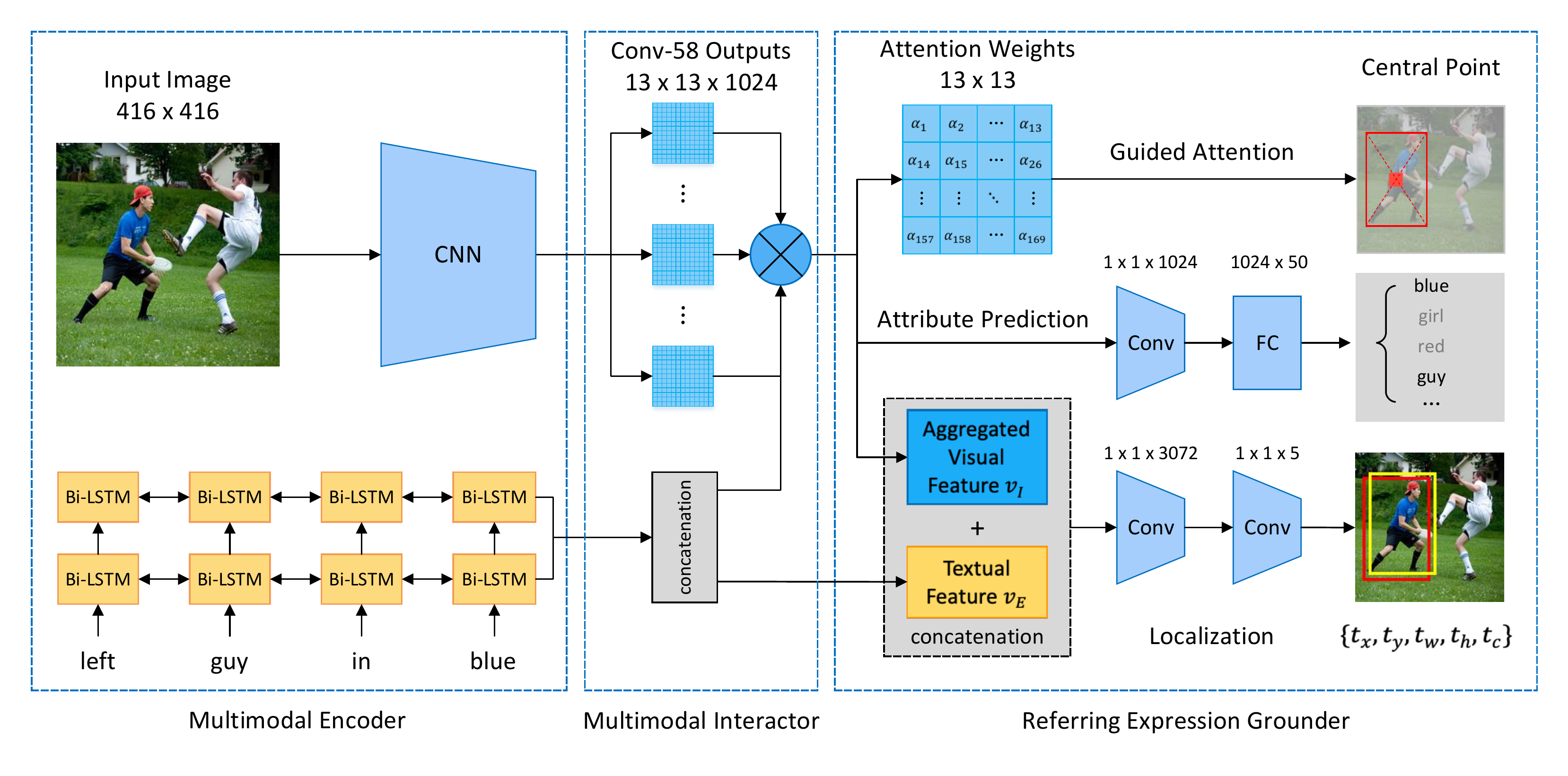}
    \vspace{-20pt}
    \caption{An overview of our proposed SSG model. The input image is encoded by a CNN to generate the local visual features representing different regions. An RNN encoder realized by a two-layer bidirectional LSTM (Bi-LSTM) is employed to process the referring expression sequentially and yield the textual feature. The multimodal interactor attentively exploits and summarizes the complicated relationships between the visual and textual features. In the referring expression grounder, the localization module relies on the joint context representations to yield the coordinates and the confidence score of the bounding box. Moreover, a novel guided attention mechanism by relating the attention weights to the referring region, enforces the visual attention to focus on the central region of the referent. Furthermore, the attribute prediction module is introduced to reproduce the attribute information contained in the referring expression. Please note that we only use the localization module to generate the bounding box for the referring expression during the inference stage.}
    \label{fig_framework}
    \vspace{-10pt}
\end{figure*}

\section{Architecture}
Given an image $I$ and a referring expression $E = \{ e_t \}^{T}_{t=1}$, where $e_t$ is the $t$-th word and $T$ denotes the total number of words, the goal of referring expression comprehension is to localize one sub-region $I_b$ within the image $I$, which corresponds to the semantic meaning of the referring expression $E$.

We propose a novel model free of region proposals, namely SSG, to tackle the referring expression comprehension task. As illustrated in Fig.~\ref{fig_framework}, our proposed SSG is a single-stage model and consists of three components. More specifically, the multimodal encoders generate the visual and textual representations for the image and referring expression, respectively. Afterward, the multimodal interactor performs a visual attention mechanism which aims to generate an aggregated visual vector by focusing on the useful region of the input image. Finally, the referring expression grounder performs the localization to predict the bounding box corresponding to the referring expression.

\subsection{Multimodal Encoder}
\label{sec:encoder}
The multimodal encoder in our SSG is used to generate the semantic representation of the input data, \emph{i.e.}, both image and text, as shown in Fig.~\ref{fig_framework}. 
\subsubsection{Image Encoder} 
\label{imageencoder}
We take an advanced CNN architecture --- YOLO-v3\footnote{https://pjreddie.com/media/files/yolov3.weights}~\cite{yolo3} --- pretrained on the MSCOCO-LOC dataset~\cite{MSCOCO} as the image encoder. Specifically, we first resize the given image $I$ to the size as $3 \times 416\times 416$, and then feed it into the encoder network. The output vectors $s=\{ \mathbf{s}_n \}^{N}_{n=1}, \mathbf{s}_{n} \in \mathbb{R}^{D_I},$  from the $58$-th convolutional layer are used as the feature representations which denote different local regions for the image. According to the network structure of YOLO-v3, $\mathbf{s}_n$ is a vector with dimension size $D_I=1024$, and the total number of local regions $N=169$.



\subsubsection{Text Encoder}
\label{textencoder}

Given a referring expression $E = \{ e_t \}^{T}_{t=1}$, where $e_t$ denotes the $t$-th word. First, each word in the referring expression needs to be initialized by the recent advanced word embedding models, such as Word2Vec~\cite{word2vec}, GloVe~\cite{glove}, and ELMo~\cite{ELMo}. In this paper, we take the EMLo model pre-trained on a dataset of 5.5B tokens to generate the corresponding word embedding vectors $w = \{ \mathbf{w}_{t} \}^{T}_{t=1}, \mathbf{w}_t \in  \mathbb{R}^{D_{w}}$, where the dimension size is $D_{w}=3072$. Afterwards, each word embedding vector $\mathbf{w}_t$ of the referring expression is fed into an RNN encoder sequentially to generate a fixed-length semantic vector as its textual feature. 

In order to adequately capture long-term dependencies between words, Long Short-Term Memory (LSTM)~\cite{Schmidhuber_1997} with specifically designed gating mechanisms is employed as the RNN unit to encode the referring expression. Moreover, the bidirectional LSTM~(Bi-LSTM)~\cite{bilstm,yu2018mattnet} can  capture the past and future context information for the referring expression, which thereby outperforms both traditional LSTMs and RNNs. In this paper, the text encoder is realized by stacking two Bi-LSTM layers together, with hidden size being $H=512$ and initial hidden and cell states setting to zeros. The semantic representation of the reference expression  is thus obtained by concatenating the forward and  backward outputs of the two stacked layers:
\begin{equation}
\small
  \mathbf{v}_E = [{\mathbf{h}}^{(1, fw)}_{T}; {\mathbf{h}}^{(1, bw)}_{T}; {\mathbf{h}}^{(2, fw)}_{T}; {\mathbf{h}}^{(2, bw)}_{T}],
\end{equation}
where ${\mathbf{h}}^{(1, fw)}_{T}$ and ${\mathbf{h}}^{(2, fw)}_{T}$ indicate the forward outputs of the first and second layers of Bi-LSTM, respectively. And ${\mathbf{h}}^{(1, bw)}_{T}$ and ${\mathbf{h}}^{(2, bw)}_{T}$ indicate the corresponding backward outputs of the first and second layers of Bi-LSTM. $\mathbf{v}_E \in \mathbb{R}^{D_E}$, with the dimension size being $D_E=2048$, denotes the finally obtained textual feature.

\subsection{Multimodal Interactor}
\label{sec:interactor}
Based on the local visual features $s$ and textual feature $\mathbf{v}_{E}$, a multimodal teractor is proposed to attentively exploit and summarize their complicated relationships. Specifically, we take the attention mechanism~\cite{XuICML2015} to aggregate the visual local features $s=\{ \mathbf{s}_n \}^{N}_{n=1}, \mathbf{s}_n \in  \mathbb{R}^{D_I}$ and generate the aggreagted visual feature $\mathbf{v}_I \in \mathbb{R}^{D_I}$ conditioned on the textual feature $\mathbf{v}_{E} \in \mathbb{R}^{D_E}$ of the referring expression:
\begin{equation}
  \small
  \mathbf{v}_{I} = f_{att}(s, \mathbf{v}_E) = \sum_{i=1}^{|s|}{\frac{\exp\left(\alpha(\mathbf{s}_i, \mathbf{v}_{E})\right)}{\sum\nolimits_{j=1}^{|s|}{\exp\left(\alpha(\mathbf{s}_{j}, \mathbf{v}_{E})\right)}} \mathbf{s}_{i}},
  \label{eq:visual_attention}
\end{equation}
where $f_{att}$ denotes the attention mechanism. $\alpha(\mathbf{s}_i, \mathbf{h}_{T})$ determines the attentive weight for the $i$-th visual local feature $\mathbf{s}_i$ with regard to the expression representation  $\mathbf{v}_{E}$, which is realized by a Multi-Layer Perceptron~(MLP):
\begin{equation}
 \small
 \alpha(\mathbf{s}_i, \mathbf{v}_{E}) = \mathbf{W}_{s_i, v_E} \text{tanh}(\mathbf{W}_{s_i}\mathbf{s}_{i} + \mathbf{W}_{v_E}\mathbf{v}_{E}),
\end{equation}
where $\mathbf{W}_{s_i, v_E} \in \mathbb{R}^{H \times N}$, $\mathbf{W}_{s_i} \in \mathbb{R}^{D_I \times H}$, and $\mathbf{W}_{v_E} \in \mathbb{R}^{D_E \times H}$ are the trainable parameters of the MLP. 

Such an attention mechanism enables each local visual feature to meet and interact with the referring expression representation, therefore attentively summarizing the visual local features together and yielding the aggregated visual context feature. Finally, by concatenating the aggregated visual context feature and the textual feature together, we can obtain the joint representation $\mathbf{v}_{I,E} \in \mathbb{R}^{D_{I, E}}$ for the image and referring expression:
\begin{equation}
  \small
  \mathbf{v}_{I, E}=[\mathbf{v}_I ; \mathbf{v}_E],
\end{equation}
where the dimension size is $D_{I,E}=3072$. Based on $\mathbf{v}_{I,E}$, our proposed referring expression grounder is proposed to localize the image region for the referring expression.

\textbf{Discussion.} \hphantom{} Note that our multimodal interactor is different from the maximum attention module proposed in GroundR~\cite{GroundR}. The local regions for attention in GroundR are first extracted by Selective Search~\cite{selectivesearch} or EdgeBoxes~\cite{edgeboxes}, and then encoded by the VGG~\cite{vgg} model. Moreover, the ``in-box" attention module proposed in~\cite{yu2018mattnet} is used to localize the relevant region within a region proposal without any auxiliary guided attention loss~(Sec.~\ref{sec:guided-attention}).

\subsection{Referring Expression Grounder}
\label{sec:grounder}
As illustrated in Fig.~\ref{fig_framework}, the referring expression grounder consists of three modules, namely localization, guided attention, and attribute prediction. We first introduce the localization module for predicting the bounding box as well as the confidence score, which relies on the coordinate information of the ground-truth referents for training. Subsequently, we introduce the guided attention mechanism and the attribute prediction modules to further improve the localization accuracy by exploiting the hidden information contained in the image as well as the referring expression.

\subsubsection{Localization} 
\label{sec:localization}
We rely on the joint representation $\mathbf{v}_{I,E}$ to predict the referring region within the image $I$, indicated by a bounding box $b_{pred}$, which semantically corresponds to the referring expression $E$. As illustrated in Fig.~\ref{fig_framework}, the joint representation $\mathbf{v}_{I,E}$ undergoes one convolutional layer with $3072$ filters and stride $1 \times 1$. Afterwards, another convolutional layer with $5$ filters and stride $1 \times 1$ followed by a sigmoid function is stacked to predict the coordinate information, which consists of $4$ values $\{ t_x, t_y, t_w, t_h \}$ and the confidence score $t_{c}$ for the predicted bounding box $b_{pred}$. Here, a convolutional layer consists of a convolution operation and an activation process, specifically the Leaky ReLU~\cite{leakyrelu}. 

\textbf{Coordinates.} The four coordinates are real values between $0$ and $1$ relative to the width and height of the image. More specifically, $t_x$ and $t_y$ denote the top-left coordinates, while $t_w$ and $t_h$ indicate the width and height of the bounding box. In order to reflect that small deviations in large bounding boxes matter less than those in small boxes, similar to~\cite{yolo}, we predict the square root of the bounding box width and height instead of the actual width and height. As such, the coordinates of the predicted bounding box are computed:
\begin{equation}
  \small
  \begin{split}
	b_{x} = t_{x} \ast p_{w}, \qquad b_{y} = t_{y} \ast p_{h}, \\
	b_{w} = t_{w}^{2} \ast p_{w}, \qquad  b_{h} = t_{h}^{2} \ast p_{h},
  \end{split}
\end{equation}
where $p_{w}$ and $p_{h}$ represent the width and the height of the input image, respectively. $\{ b_x, b_y \}$, $b_w$, and $b_h$ denote the top-left coordinates, width, and height of the predicted bounding box $b_{pred}$, respectively. During the training, the mean squared error~(MSE) is used as the objective function:
\begin{equation}
  \small
  \begin{aligned}
  \mathcal{L}_{\text{loc}} &= \left( t_x - \frac{\hat{b}_{x}}{p_w} \right)^2 + \left( t_y - \frac{\hat{b}_{y}}{p_h} \right)^2 \\
   & \qquad + \left( t_w - \sqrt{\frac{\hat{b}_{w}}{p_w}} \right)^2 + \left( t_h - \sqrt{\frac{\hat{b}_{h}}{p_h}} \right)^2,
  \end{aligned}
  \label{eq:localization}
\end{equation}
where $\hat{b}_{x}, \hat{b}_{y}, \hat{b}_{w}, \hat{b}_{h}$ are the coordinate information of the ground-truth bounding box $b_{gt}$.

\textbf{Confidence Score.} \hphantom{} As aforementioned, besides the coordinate information, the localization module will also generate a confidence score $t_{c}$, reflecting the accuracy of the predicted box. During the evaluation, a predicted bounding box is regarded as a correct comprehension result if the intersection-over-union~(IoU) of the box with the ground-truth bounding box is larger than a threshold $\eta$. Usually, the threshold is set to $\eta=0.5$. Therefore, we naturally realize the confidence score prediction as a binary classification problem rather than a regression problem as YOLO~\cite{yolo}. Hence the target confidence score $\hat{b}_{c}$ is defined as:
\begin{equation}
  \small
  \hat{b}_{c} = \begin{cases}
                    1, & \text{if } {IoU} \left( b_{pred}, b_{gt} \right) \ge \eta\\
                    0, & \text{otherwise}
                   \end{cases}
\end{equation}
The objective function regarding the confidence score is defined as a binary cross-entropy:
\begin{equation}
  \small
  \mathcal{L}_{\text{conf}} = \hat{b}_{c} * \text{log}(t_{c}) + (1 - \hat{b}_{c}) * \text{log}(1 - t_{c}).
\end{equation}	
Please note that the objective function regarding confidence score is different from the definition in~\cite{yolo,yolo2}, which is considered as a regression problem and formulated as $Pr(b_{gt}) * IoU(b_{pred}, b_{gt})$, where $Pr(b_{gt})$ is equal to 1 when there is an object in the cell, and 0 otherwise.


\begin{figure}[!t]
    \centering
    \includegraphics[width=\linewidth]{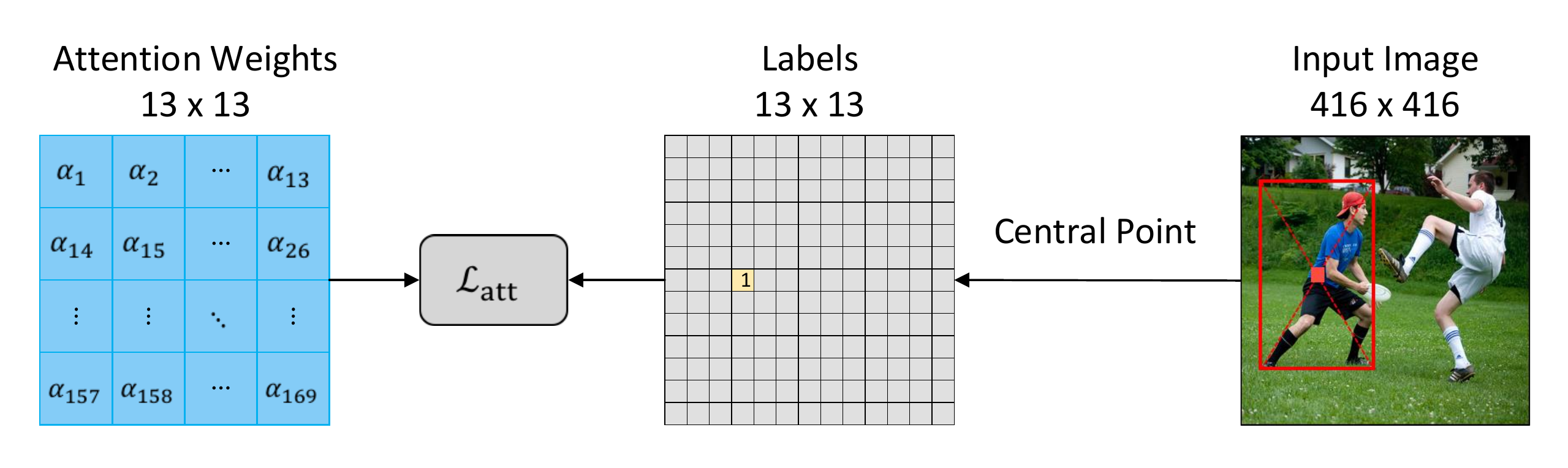}
    \caption{The illustration of our proposed guided attention loss. We formulate the guided attention process as a classification problem with the local region, where the central point falls into being labeled as 1 and the rest labeled as 0.} 
    \label{fig_att}
    \vspace{-10pt}
\end{figure}

\subsubsection{Guided Attention} 
\label{sec:guided-attention}
For further boosting the grounding accuracy, we propose a guided attention mechanism to encourage our model to pay more attention to the central region of the correct referent. As introduced in Sec.~\ref{sec:interactor}, a set of attention weights $\mathbf{\alpha} = \{ {\alpha}_{n} \}^{N}_{n=1}, {\alpha}_{n} \in \mathbb{R}$ are computed conditioned on the textual feature for different visual local features, with each representing its relevance to the referring expression. We notice that there exists one piece of hidden information, namely \textit{object center bias}~\cite{centerbias}, which we can make full use of. The central region of the ground-truth bounding box should produce the maximum attention weight since the visual feature related to the central region is more important for grounding the referring expression. To this end, as illustrated in Fig.~\ref{fig_att}, we first identify the position of the center point using the ground-truth bounding box, and encode it into a one-hot vector as the label $\hat{\mathbf{y}}$, which means that only the region cell, where the central point of the referent falls into, is labeled as 1 with all the rest labeled as 0. The coordinates of the central point after rescaling to the size of the attention weight map are given by: 
\begin{equation}
  \small
  \Bigg(\left\lfloor \frac{\hat{b}_x + 0.5 \times \hat{b}_w}{m} \right\rfloor, \left\lfloor \frac{\hat{b}_y + 0.5 \times \hat{b}_h}{m} \right\rfloor\Bigg).
\end{equation}
As mentioned in Sec.~\ref{imageencoder}, the sizes of the attention weight map and the input image are $13 \times 13$ and $416 \times 416$, respectively. Therefore, the rescaling factor $m$ is set to $416/13=32$. Finally, we use the cross-entropy loss as the objective function to measure the difference between the visual attention weights and the obtained one-hot label $\hat{\mathbf{y}}$:
\begin{equation}
  \small
  \mathcal{L}_{\text{att}} = - \sum_{i}^{N}{\hat{y}_{i}\text{log}\alpha_{i}},
\end{equation}
where $\hat{y}_{i}$ denotes the $i$-th entry of the label vector $\hat{\mathbf{y}}$. $N$ denotes the number of attention weights, which is equal to $13\times 13 = 169$. Such auxiliary loss can help our model learn to discriminate the target region with the other ones and encourage the attentive visual feature to embed more important information for predicting the bounding box.

\begin{table*}[!t]
  \begin{center}
  \small
  \tabcolsep=0.32cm
  \caption{The performance comparisons (Acc$\%$) of different methods on RefCOCO, RefCOCO+, and RefCOCOg datasets. The best results among all models are marked with boldface.}
  \label{tb_comparison}
   \begin{tabular}{|c|l|cccccc|}
      \hline
       \multirow{2}{*}{Line} & \multirow{2}{*}{Models} & \multicolumn{2}{c}{RefCOCO} & \multicolumn{2}{c}{RefCOCO+} & RefCOCOg & RefCOCOg \\
        &  & test~A & test~B & test~A & test~B & val~(\textit{google}) & val~(\textit{umd}) \\
      \hline
      \hline
      1 & MMI~\cite{mao_MMI} & 64.90 & 54.51 & 54.03 & 42.81 & 45.85 & - \\
      2 & Vis-Diff + MMI~\cite{yu2016modeling} & 67.64 & 55.16 & 55.81 & 43.43 & 46.86 & - \\
      3 & Neg-Bag~\cite{nagaraja16refexp} & 58.70 & 56.40 & - & - & - & 49.50 \\
      4 & Attr + MMI + Vis-Diff~\cite{liu_iccv17_attr} & 72.08 & 57.29 & 57.97 & 46.20 & 52.35 & - \\
      5 & CMN~\cite{hu_cvpr2017_cmn} & 71.03 & 65.77 & 54.32 & 47.76 & 57.47 & -  \\
      6 & Speaker + Listener + MMI~\cite{yu_cvpr2017_joint} & 72.95 & 62.43 & 58.58 & 48.44 & 57.34 & - \\
      7 & Speaker + Listener + Reinforcer + MMI~\cite{yu_cvpr2017_joint} & 72.94 & 62.98 & 58.68 & 47.68 & 57.72 & - \\
      8 & Variational Context~\cite{zhang2017grounding} & 73.33 & 67.44 & 58.40 & 53.18 & \textbf{62.30} & - \\
      9 & MAttNet~\cite{yu2018mattnet} & \textbf{80.43} & \textbf{69.28} & \textbf{70.26} & \textbf{56.00} & - & \textbf{66.67} \\
      \hline
      \hline
      10 & SSG ($\lambda_{\text{loc}}$) & 72.90 & 63.97 & 23.00 & 16.51 & 17.64 & 18.83 \\
      11 & SSG ($\lambda_{\text{loc} + \text{conf}}$) & 73.44 & 64.39 & 58.16 & 43.55 & 42.10 & 51.97 \\
      12 & SSG ($\lambda_{\text{loc} + \text{conf} + \text{att}}$) & 75.20 & 65.77 & 61.39 & 46.50 & 43.90 & 56.63 \\
      13 & SSG ($\lambda_{\text{loc} + \text{conf} + \text{att} + \text{attr}}$) & 76.51 & 67.50 & 62.14 & 49.27 & 47.78 & 58.80 \\
      \hline
    \end{tabular}
    \vspace{-15pt}
  \end{center}
\end{table*}

\subsubsection{Attribute Prediction}
\label{sec:attribute-prediction}
Additionally, visual attributes are usually used to distinguish referent objects of the same category and have shown impressive performance on many multimodal tasks, such as image captioning~\cite{ltg,yao2017boosting,you2016image}, video captioning~\cite{pan2017video}, and referring expression comprehension~\cite{liu_iccv17_attr,yu2018mattnet}. Inspired by the previous work~\cite{yu2018mattnet}, we introduce an attribute prediction module to further boost the performance of our grounder. As illustrated in Fig.~\ref{fig_framework}, the attentively aggregated visual feature $\mathbf{v}_{I}$ undergoes an additional convolutional layer with 1024 filters and stride $1 \times 1$. A fully connected layer is subsequently stacked to predict the probabilities $\{ p_i \}^{N_{attr}}_{i=1}$ for all $N_{attr}$ attributes, where $N_{attr}$ is the number of the most frequent attribute words extracted from the training dataset\footnote{https://github.com/lichengunc/refer-parser2}. In this paper, we empirically set $N_{attr}=50$ as~\cite{yu2016modeling}. As such, the attribute prediction can be formulated as a multi-label classification problem, whose objective function is defined as: 
\begin{equation}
  \small
  \mathcal{L}_{\text{attr}} = \sum_{i=1}^{N_{attr}} w^{attr}_{i} \left( \hat{y}_{i} \text{log}(p_{i}) + (1 - \hat{y}_{i}) \text{log}(1 - p_{i}) \right),
\end{equation}
where $w^{attr}_{i}=1/\sqrt{\text{freq}_{attr}}$ is used to balance the weights of different attributes. $\hat{y}_{i}$ is set to 1 when the $i$-th attribute word exists in the referring expression, and 0 otherwise. During training, the loss value of attribute prediction is set to zero if there is no attribute word existing in the referring expression.



\subsection{Training Objective}
\label{sec:training}
The objective function of our SSG model for a single training sample \textit{(image, referring expression, bounding box)} is defined as a weighted sum of the aforementioned localization loss, the confidence score loss, the guided attention loss, and the attribute prediction loss:
\begin{equation}
  \small
  \mathcal{L}_{\text{sum}} = \lambda_{\text{loc}} \mathcal{L}_{\text{\text{loc}}} + {\lambda_{\text{conf}} \mathcal{L}_{\text{conf}}} + \lambda_{\text{att}} \mathcal{L}_{\text{att}} + \lambda_{\text{attr}} \mathcal{L}_{\text{attr}},
  \label{eq:training_objective}
\end{equation}
where $\lambda_{\text{loc}}$, $\lambda_{\text{conf}}$, $\lambda_{\text{att}}$, and $\lambda_{\text{attr}}$ are the weight factors to balance the contributions of different losses for model training. 

\subsection{Inference}
\label{sec:infernce}
During the inference phase, only the localization module is enabled to predict the bounding box, which corresponds to the referring expression, with the guided attention and attribute prediction modules deactivated. For one given image $I$ and the corresponding referring expression $E$, these modules, namely the multimodal encoder (including image and text encoders), multimodal interactor, and localization, fully couple with each other and accordingly predict the bounding box $b_{pred}$ in one single stage. As such, our SSG performs more efficiently for referring expression comprehension compared with the existing multi-stage models, which will be further demonstrated in Sec.~\ref{sec:efficiency}.

\section{Experiments}
\label{sec:experiments}
\subsection{Datasets}
We evaluate and compare our proposed SSG with existing approaches comprehensively on the four popular datasets, namely RefCOCO~\cite{yu2016modeling}, RefCOCO+~\cite{yu2016modeling}, RefCOCOg~\cite{mao_MMI}, and ReferItGame~\cite{referitgame}.

RefCOCO, RefCOCO+, and RefCOCOg were all collected from the MSCOCO~\cite{MSCOCO} dataset, but with several differences. \textbf{(1).}~The expressions in RefCOCO contain many location words (\textit{e.g.} ``left", ``corner"). While RefCOCO+ was collected to encourage the expressions to focus on the appearance of the referent without using location words. RefCOCOg contains longer referring expressions on average than RefCOCO and RefCOCO+ (8.4 \textit{vs.} 3.5) and provides more embellished expressions than RefCOCO and RefCOCOg. \textbf{(2).}~Both RefCOCO and RefCOCO+ are divided into train, validation, test~A containing person referents, and test~B containing common object referents. While RefCOCOg has two types of data partitions. The first split is denoted as \textit{google} which was used in \cite{mao_MMI}. Since the testing set has not been released, recent work~\cite{hu_cvpr2017_cmn,liu_iccv17_attr,mao_MMI,yu2018mattnet,yu2016modeling,yu_cvpr2017_joint,zhang2017grounding} reported their results on the validation set. The second split is denoted as \textit{umd} which was used in \cite{nagaraja16refexp,yu2018mattnet}. In this paper, we evaluate our model on both types of data splits for RefCOCOg.

ReferItGame also named as RefCLEF was collected from the segmented and annotated extension of the ImageCLEF IAPR TC-12 dataset~(SAIAPR TC-12)~\cite{saiapr}. Note that the annotated expressions provided by this dataset exist some equivocal words and erroneous annotations, such as \textit{anywhere} and \textit{don't know}. In this paper, we use the same data split as \cite{hu_cvpr2017_cmn,SCRC,GroundR,zhang2017grounding} for fair comparison.

\begin{table}[!t]
  \begin{center}
  \tabcolsep=0.18cm
  \small
  \caption{The performance comparisons (Acc\%) of different methods on the ReferItGame dataset.}
  \label{tb_comparison2}
    \begin{tabular}{|c|l|c|c|}
      \hline
      Line & Models & Proposal & ReferItGame \\
      \hline
      \hline
     1 & SCRC~\cite{SCRC} & \multirow{6}{*}{EdgeBoxes} & 17.93 \\
     2 & GroundR~\cite{GroundR} &  & 26.93 \\
     3 & CMN~\cite{hu_cvpr2017_cmn} &  & 28.33 \\
     4 & Variational Context~\cite{zhang2017grounding} &  & 31.13 \\
     5 & MAttNet &  & 29.04 \\
     6 & Oracle &  & 59.45 \\
     \hline
     \hline
     7 & SSG ($\lambda_{\text{loc}}$) & \multirow{4}{*}{---} & 49.68 \\
     8 & SSG ($\lambda_{\text{loc} + \text{conf}}$) &  & 49.97 \\
     9 & SSG ($\lambda_{\text{loc} + \text{conf} + \text{att}}$) &  & 54.14 \\
     10 & SSG ($\lambda_{\text{loc} + \text{conf} + \text{att} + \text{attr}}$) &  & \textbf{54.24} \\
      \hline
    \end{tabular}
    \vspace{-15pt}
  \end{center}
\end{table}

\subsection{Experiment Settings}
\textbf{Proprocessing.} \phantom{} As aforementioned, we initialize the word embedding layers in our model with EMLo~\cite{ELMo}, which is a character-based embedding model. Special characters are removed, resulting in a vocabulary size of 10,342, 12,227, 12,679, and 9,024 for RefCOCO, RefCOCO+, RefCOCOg, and ReferItGame, respectively. We truncate all the referring expressions longer than 15 words and use zero padding for the expressions shorter than 15 words.

\textbf{Training.} \phantom{} To balance the contribution of each loss for optimal model training in Eq.~\ref{eq:training_objective}, we empirically set $\lambda_{\text{loc}}$, $\lambda_{\text{conf}}$, $\lambda_{\text{att}}$ and $\lambda_{\text{attr}}$ to 20.0, 5.0, 1.0, and 5.0, respectively. The SGD optimizer with an initial learning rate of $1 \times 10^{-3}$ and the momentum setting as 0.9 is employed to train our model. The learning rate is decreased by 0.8 every 5 epochs. All the expressions for the same referent are tied into one single batch samples for training. Early stopping is used to prevent overfitting if the performance on the validation set does not improved over the last 10 epochs. Our SSG is implemented with PyTorch and can be trained within 100 hours on a single Tesla P40 and CUDA 9.0 with Intel Xeon E5-2699v4@2.2GHz.




\textbf{Evaluation Metric.} \phantom{} Same as the previous work~\cite{SCRC,nagaraja16refexp,yu2018mattnet,zhang2017grounding}, we evaluate the performance of our model using the ratio of Intersection over Union~(IoU) between the ground truth and the predicted bounding box. If the IoU is larger than 0.5, we treat this predicted bounding box as a true positive. Otherwise it is a false positive. The fraction of the true positive expressions are denoted as the final accuracy.

\begin{table}[!t]
  \centering
  \small
  \tabcolsep=0.22cm
  \caption{The inference time~(seconds per referent) comparisons on the RefCOCO dataset between our SSG, SCRC, and MAttNet. \textit{Env.} means the hardware environment.}
  \label{Table_speed}
  \begin{tabular}{|l|c|ccc|c|}
      \hline
      Models & Env. & Stage \rom{1} & Stage \rom{2} & Stage \rom{3}  & Total \\
      \hline
      \hline
       SCRC & \multirow{3}{*}{CPU} & 0.353 & 0.511 & 10.781 & 11.645 \\
       MAttNet &  & 14.907 & 0.849 & 0.157 & 15.913 \\
       SSG &  & - & - & - & \textbf{1.373} \\
      \hline
      \hline
      SCRC & \multirow{3}{*}{GPU} & 0.353 & 0.025 & 0.272 & 0.650 \\
      MAttNet &  & 0.183 & 0.043 & 0.010 & 0.236 \\
      SSG &  & - & - & - & \textbf{0.025} \\
      \hline
  \end{tabular}
  \vspace{-15pt}
\end{table}


\subsection{Performance Comparisons}

\begin{figure*}[ht]
    \centering
    \includegraphics[width=\linewidth]{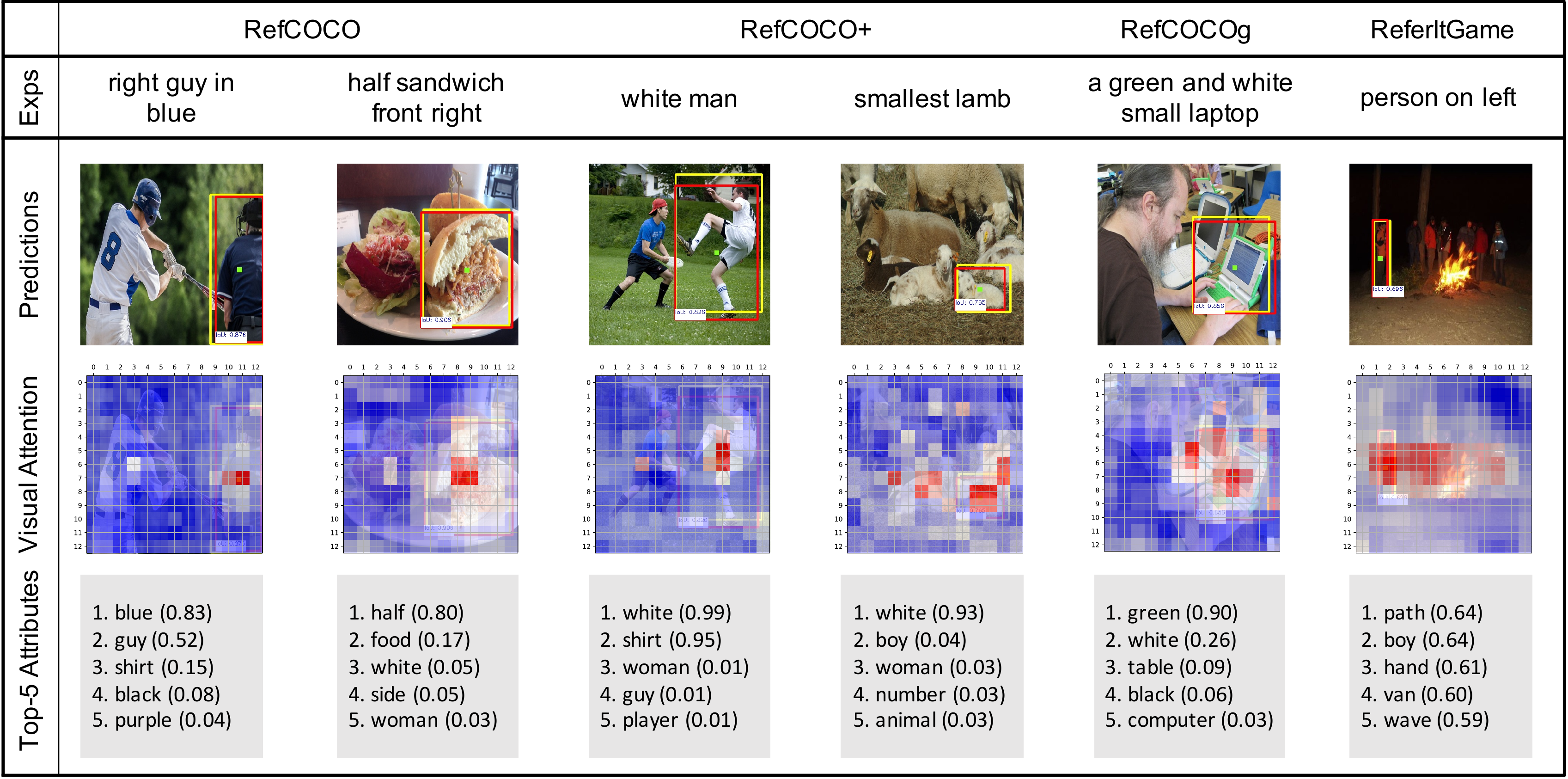}
    \caption{Qualitative results of the referring expression comprehensions with the corresponding visual attention heat maps and top-5 predicted attributes. The red rectangles denote the ground-truth bounding boxes, while the yellow ones denote the predicted boxes by our SSG. The green dots indicate the center points of the ground-truth bounding boxes.} 
    \label{fig_exp_pos}
    \vspace{-10pt}
\end{figure*}

We compare our SSG with existing multi-stage methods comprehensively. For the fair comparison, we directly copy the results from their papers.

The results on RefCOCO, RefCOCO+, and RefCOCOg are shown in Table~\ref{tb_comparison}. Although it is more challenge to localize the referent without resorting to region proposals directly, the results of our SSG~(Line 13) on the test~A and test~B split of RefCOCO outperform most of the previous models, except MAttNet~\cite{yu_cvpr2017_joint}. RefCOCO+ is more challenge than RefCOCO since the referring expressions in RefCOCO+ are annotated with appearance words without location information. Nevertheless, our SSG can take the second and third place on the test A and test B split of RefCOCO+, respectively. On the validation set of RefCOCOg split by \textit{google}, our model achieves favorable results which is better than~\cite{mao_MMI,yu2016modeling}. Furthermore, although the performance of SSG on the validation set of RefCOCOg split by \textit{umd} is worse than the best model MAttNet, it still outperforms \cite{nagaraja16refexp}. One reason may be that the language used in RefCOCOg tend to be more flowery than the expressions in RefCOCO and RefCOCO+~\cite{yu2016modeling}.


The performance comparisons on ReferItGame of different models are shown in Table.~\ref{tb_comparison2}. The upper-bound result of the region proposals extracted by EdgeBoxes~\cite{edgeboxes} is only 59.45\%~(Line 6), which is denoted by ``Oracle" as in SCRC~\cite{SCRC}. We use the released code as well as off-the-shelf proposals provided by the authors\footnote{https://github.com/lichengunc/MAttNet}~\cite{yu2018mattnet} to evaluate the performance of MAttNet on the ReferItGame dataset~(Line~5). It can be observed that our SSG outperforms all the previous models. One reason may be attributed to the low-quality proposals providing for the ReferItGame dataset, which constraint the performance of the previous multi-stage models. In contrast, the result of MAttNet evaluated by ourselves using the ground-truth bounding boxes of ReferItGame as region proposals is 81.29\%. Our model achieves the much better performance than the existing methods since it can be trained and optimized end-to-end without resorting to region proposals.
Fig.~\ref{fig_exp_pos} shows some qualitative results of referring expression comprehension using our proposed SSG, as well as the visualizations of  attention weights and top-5 predicted attributes\footnote{More qualitative results and failure cases of the referring expression comprehension can be found in Appendix.}. First, our SSG can accurately ground the referents in the images. Second, by visualizing the attention weights, we can observe that our guided attention mechanism can enforce the visual attention mechanism to focus on the meaningful region of the image. And the top-5 predicted attribute words can accurately characterize the attribute information of the referents, such as ``blue", ``white", and ``food".





\subsection{Ablation Study}
\label{sec:experiments:ablation}
We perform ablation studies to examine the contribution of each component of SSG. The results are shown in Table~\ref{tb_comparison} (Line 10 - 13) and Table~\ref{tb_comparison2} (Line 7 - 10). As a baseline, the performance of $\text{SSG}~(\lambda_{\text{loc}})$ trained with localization loss only is illustrated in Table~\ref{tb_comparison} and Table~\ref{tb_comparison2}. By incorporating the confidence score loss, the performance of $\text{SSG}~(\lambda_{\text{loc + conf}})$ can be improved obviously. The performance of $\text{SSG}~(\lambda_{\text{loc + conf + att}})$ by adding the guided attention loss can be further improved. By further introducing the attribute prediction loss, the performance of $\text{SSG}~(\lambda_{\text{loc + conf + att + attr}})$ can be boosted consistently. 

\subsection{Efficiency}
\label{sec:efficiency}
We measure the speed by calculating the average time per referent~\textit{(image, referring expression)} at inference stage on the RefCOCO dataset running on the GPU-enabled and CPU-only environments. Table~\ref{Table_speed} shows the comparisons between SSG, SCRC~\cite{SCRC}, and MAttNet~\cite{yu2018mattnet}. Please note that the computation time of EdgeBoxes\footnote{https://github.com/pdollar/edges}~\cite{edgeboxes}, SCRC\footnote{https://github.com/ronghanghu/natural-language-object-retrieval}~\cite{SCRC}, and MAttNet~\cite{yu2018mattnet} are all obtained by using the author-released code under the same hardware environment. We can observe that all the models with GPU-enabled achieve significant speedups compared with the CPU implementations. When we activate the GPU for acceleration, SCRC takes the longest time due to the computation time cost by EdgeBoxes at the proposal extraction stage. MAttNet uses Faster R-CNN~\cite{faster-rcnn} for proposal extraction and takes shorter computation time at 0.236s. However, our SSG can significantly reduce the computation time to 0.025s for a referring expression along with an image, running at 40 referents per second, which is more than 9$\times$ faster than MAttNet.

\section{Conclusion}
In this paper, we proposed a novel grounding model, namely Single-Stage Grounding network~(SSG), to directly localize the referent within the given image semantically corresponding to a referring expression without resorting to region proposals. To encourage the multimodal interactor to focus on the useful region for grounding, a guided attention loss based on the object center-bias is proposed. Furthermore, by introducing attribute prediction loss, the performance can be improved consistently. Experiments on four public datasets show that our SSG model can achieve favorable performance, especially achieving  the state-of-art performance on the ReferItGame dataset. Most importantly, our model is fast by design and able to run at 40 referents per second averagely on the RefCOCO dataset.

{
  \small
  \bibliographystyle{ieee}
  \bibliography{bibtex}
}


\onecolumn


\setcounter{section}{0}
\renewcommand\thesection{\Alph{section}}
\renewcommand\thesubsection{\thesection.\arabic{subsection}}

\section{Appendix}
\subsection{Datasets}
For the referring expression comprehension task, a number of datasets have been used in the previous work~\cite{mao_MMI,SCRC,yu2016modeling,GroundR,nagaraja16refexp,liu_iccv17_attr,hu_cvpr2017_cmn,yu_cvpr2017_joint,zhang2017grounding,yu2018mattnet}, which are summarized in Table~\ref{tb_datasets}. For comparing our SSG with the previous methods comprehensively, we take the four commonly used datasets for our experiments, which are RefCOCO, RefCOCO+, RefCOCOg, and ReferItGame. Please note that the RefCOCOg dataset has two types of data splits.

\begin{table*}[ht]
  \small
  \caption{The datasets used for referring expression comprehension.}
  \vspace{5pt}
  \tabcolsep=0.18cm
  \label{tb_datasets}
  \begin{center}
    \begin{tabular}{|l|ccccccc|}
      \hline
      Models & RefCOCO & RefCOCO+ & RefCOCOg & ReferItGame & Flickr30K & Visual Genome & Kitchen \\
      \hline
      \hline
      MMI~\cite{mao_MMI} & \checkmark &  & \checkmark &  &  &  &  \\
      SCRC~\cite{SCRC} &  &  &  & \checkmark & \checkmark &  & \checkmark \\
      VisDiff + MMI~\cite{yu2016modeling} & \checkmark & \checkmark & \checkmark &  &  &  &  \\
      GroundR~\cite{GroundR} &  &  &  & \checkmark & \checkmark &  &  \\
      Neg-Bag~\cite{nagaraja16refexp} & \checkmark &  & \checkmark &  &  &  &  \\
      Attribute + VisDiff~\cite{liu_iccv17_attr} & \checkmark & \checkmark & \checkmark &  &  &  &  \\
      CMN~\cite{hu_cvpr2017_cmn} &  &  & \checkmark &  &  & \checkmark &  \\
      Speaker-Listener-Reinforcer~\cite{yu_cvpr2017_joint} & \checkmark & \checkmark & \checkmark &  &  &  &  \\
      Variational Context~\cite{zhang2017grounding} & \checkmark & \checkmark & \checkmark & \checkmark &  &  &  \\
      MAttNet ~\cite{yu2018mattnet} & \checkmark & \checkmark & \checkmark &  &  &  &  \\
      \hline
      \hline
      Our SSG & \checkmark & \checkmark & \checkmark & \checkmark &  &  &  \\
      \hline
    \end{tabular}
  \end{center}
\end{table*}

\subsection{Effect of End-to-end Training}
Our proposed SSG is an end-to-end model. The parameters in all components can be optimized jointly by stochastic gradient descent methods. As illustrated in Table~\ref{tb_ablation}, we report the results when freezing the parameters of the image encoder and compare them to the results with the fine-tuning strategy. The performance can be consistently improved by fine-tuning the image encoder, demonstrating the advantage of the end-to-end training strategy.

\begin{table*}[ht]
  \small
  \centering
  \tabcolsep=0.29cm
  \caption{Ablation study of SSG with and without fine-tuning strategy on the four datasets, which are RefCOCO, RefCOCO+, RefCOCOg, and ReferItGame.}
  \vspace{5pt}
  \label{tb_ablation}
  \begin{tabular}{|l|c|ccccccc|}
      \hline
      \multirow{2}{*}{Models} & \multirow{2}{*}{Fine-tuning}  & \multicolumn{2}{c}{RefCOCO} & \multicolumn{2}{c}{RefCOCO+} & RefCOCOg & RefCOCOg & ReferItGame \\
                              &  & test A & test B & test A & test B & val~(\textit{google}) & val~(\textit{umd}) & test \\
      \hline
      \hline
      SSG ($\lambda_{\text{loc} + \text{conf} + \text{att} + \text{attr}}$) & No & 52.88 & 48.26 & 35.88 & 30.36 & 33.25 & 37.00 & 40.05 \\
      SSG ($\lambda_{\text{loc} + \text{conf} + \text{att} + \text{attr}}$) & Yes & 76.51 & 67.50 & 62.14 & 49.27 & 47.78 & 58.80 & 54.24 \\
      \hline
  \end{tabular}
\end{table*}

\clearpage
\subsection{More Examples}
\begin{figure}[ht]
  \centering
  \includegraphics[width=\linewidth]{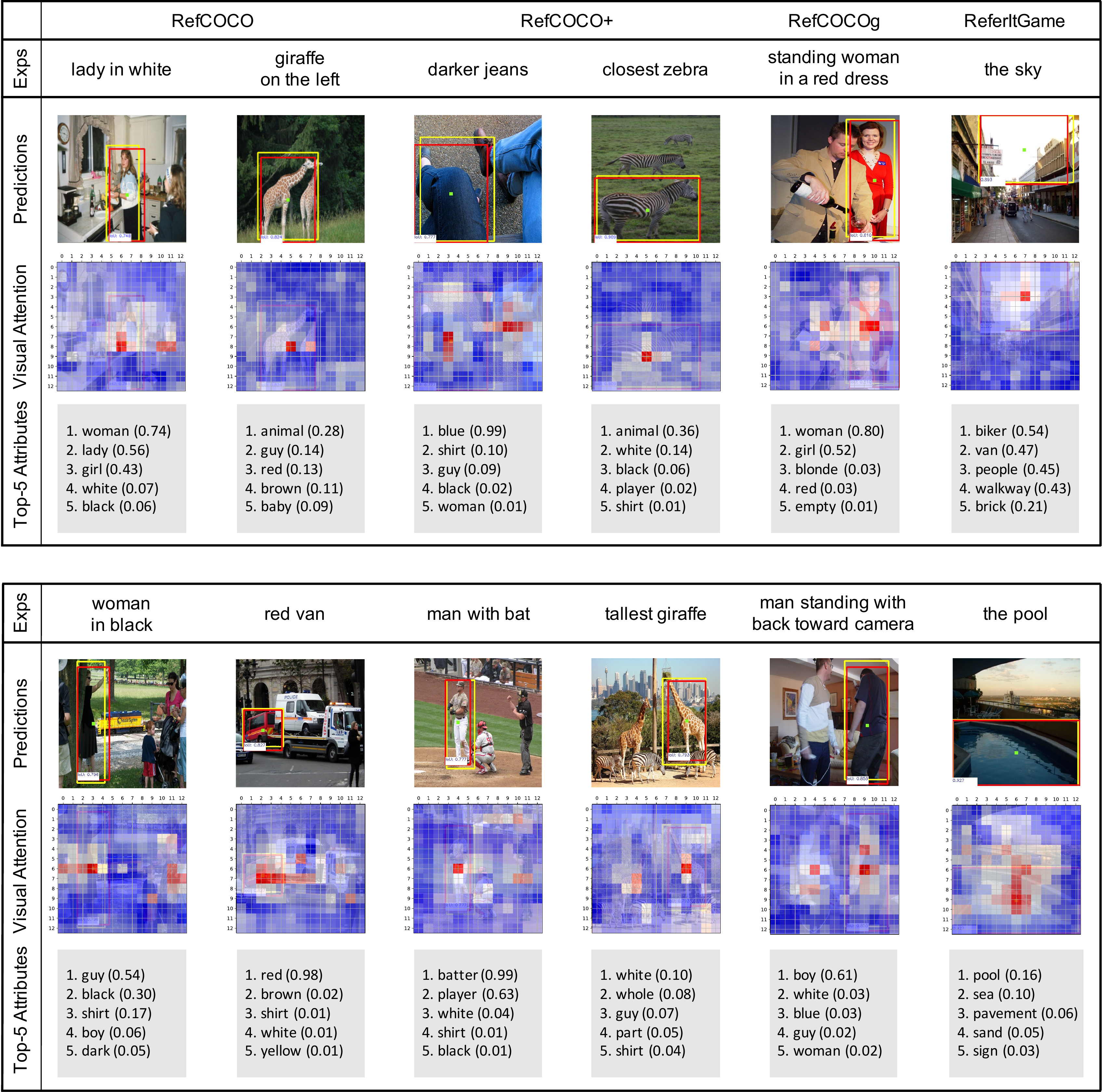}
  \caption{Qualitative results of the referring expression comprehensions with the corresponding visual attention heat maps and top-5 predicted attributes. The red rectangles denote the ground-truth bounding boxes, while the yellow ones denote the predicted boxes by our SSG. The green dots indicate the center points of the ground-truth bounding boxes.}
  \label{fig_qua1}
\end{figure}

\begin{figure}[th]
  \centering
  \includegraphics[width=\linewidth]{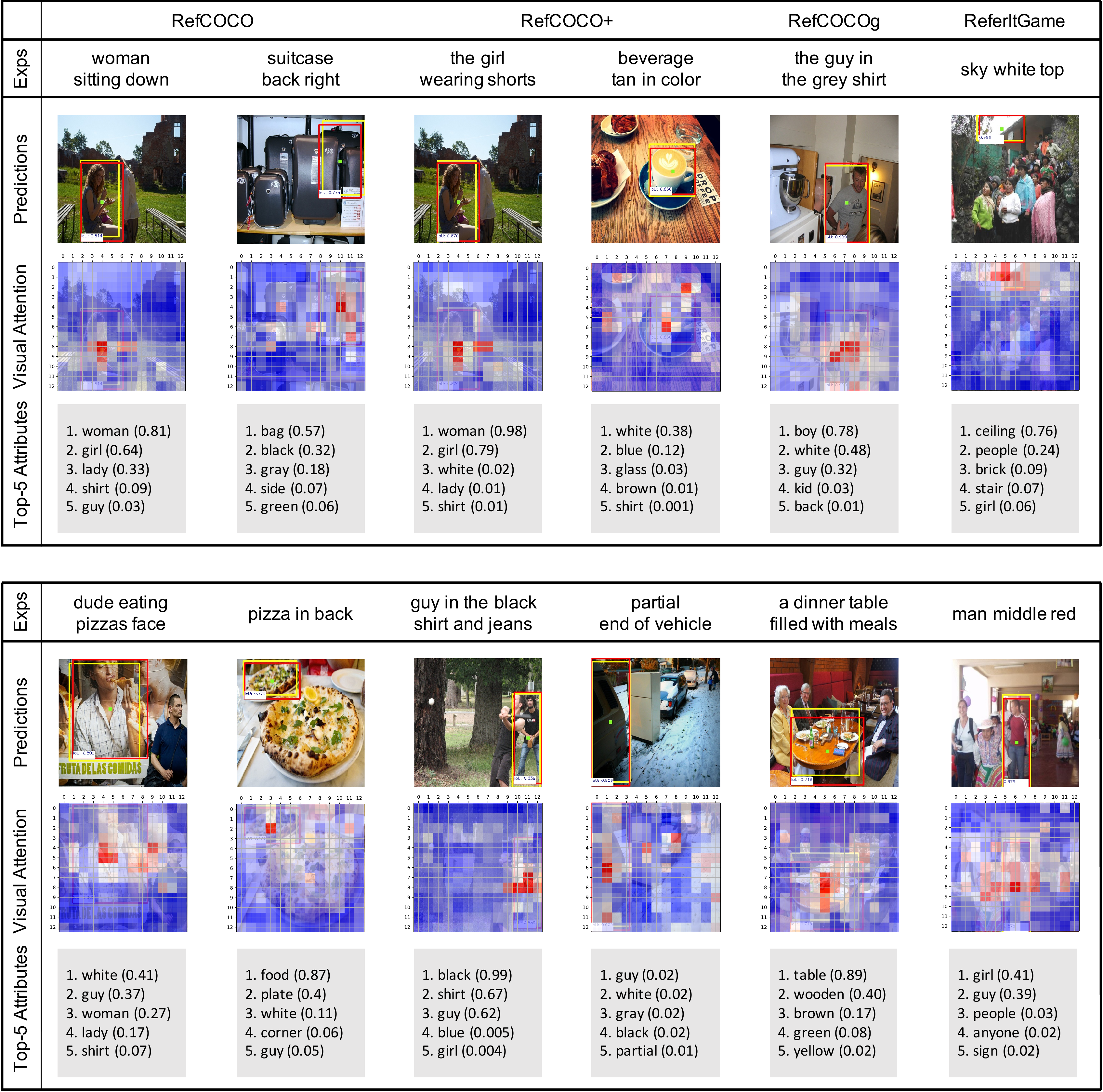}
  \caption{Qualitative results of the referring expression comprehensions with the corresponding visual attention heat maps and top-5 predicted attributes. The red rectangles denote the ground-truth bounding boxes, while the yellow ones denote the predicted boxes by our SSG. The green dots indicate the center points of the ground-truth bounding boxes.}
  \label{fig_qua2}
\end{figure}

\begin{figure}[ht]
  \centering
  \includegraphics[width=\linewidth]{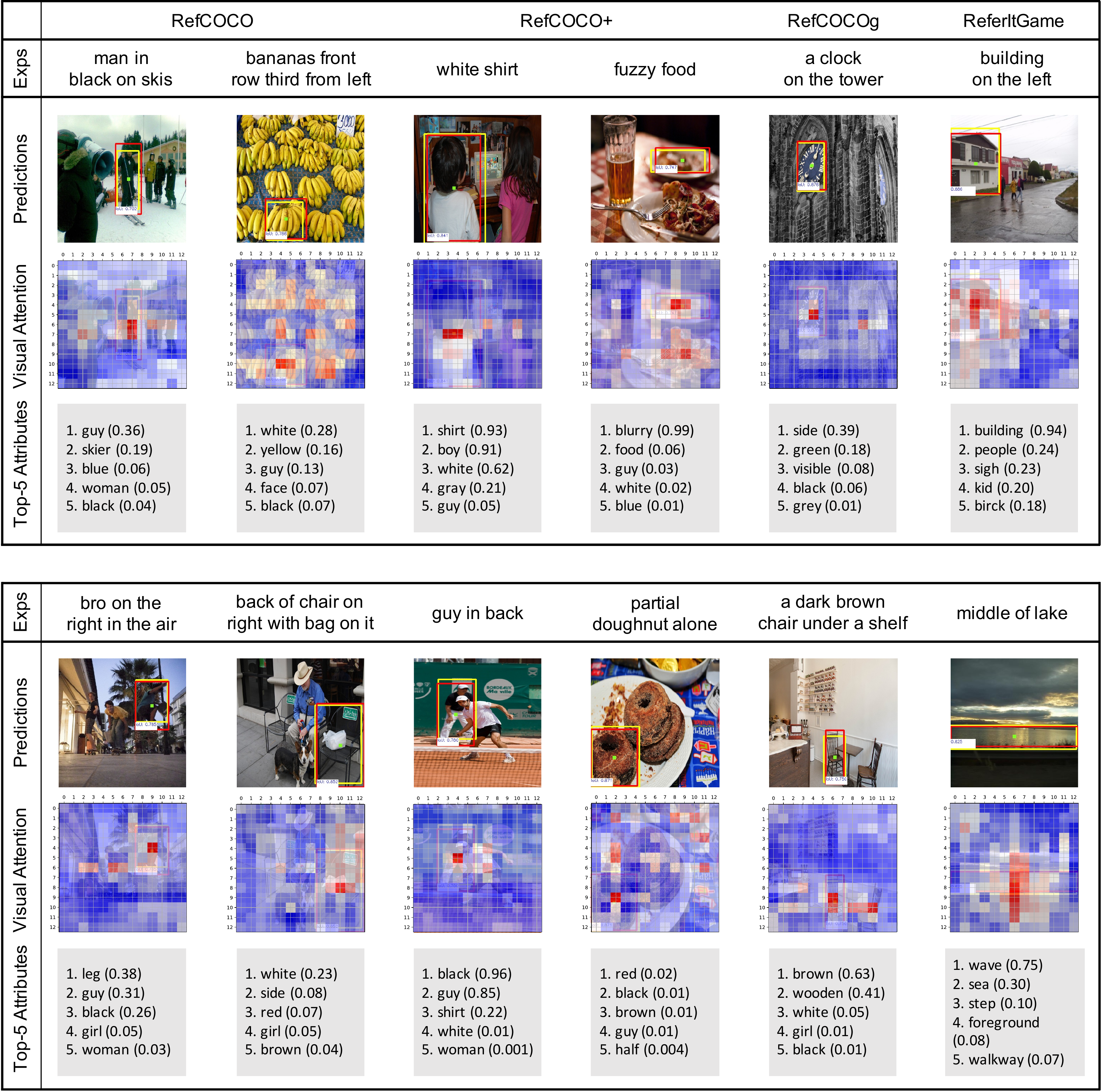}
  \caption{Qualitative results of the referring expression comprehensions with the corresponding visual attention heat maps and top-5 predicted attributes. The red rectangles denote the ground-truth bounding boxes, while the yellow ones denote the predicted boxes by our SSG. The green dots indicate the center points of the ground-truth bounding boxes.}
  \label{fig_qua3}
\end{figure}

\begin{figure}[ht]
  \centering
  \includegraphics[width=\linewidth]{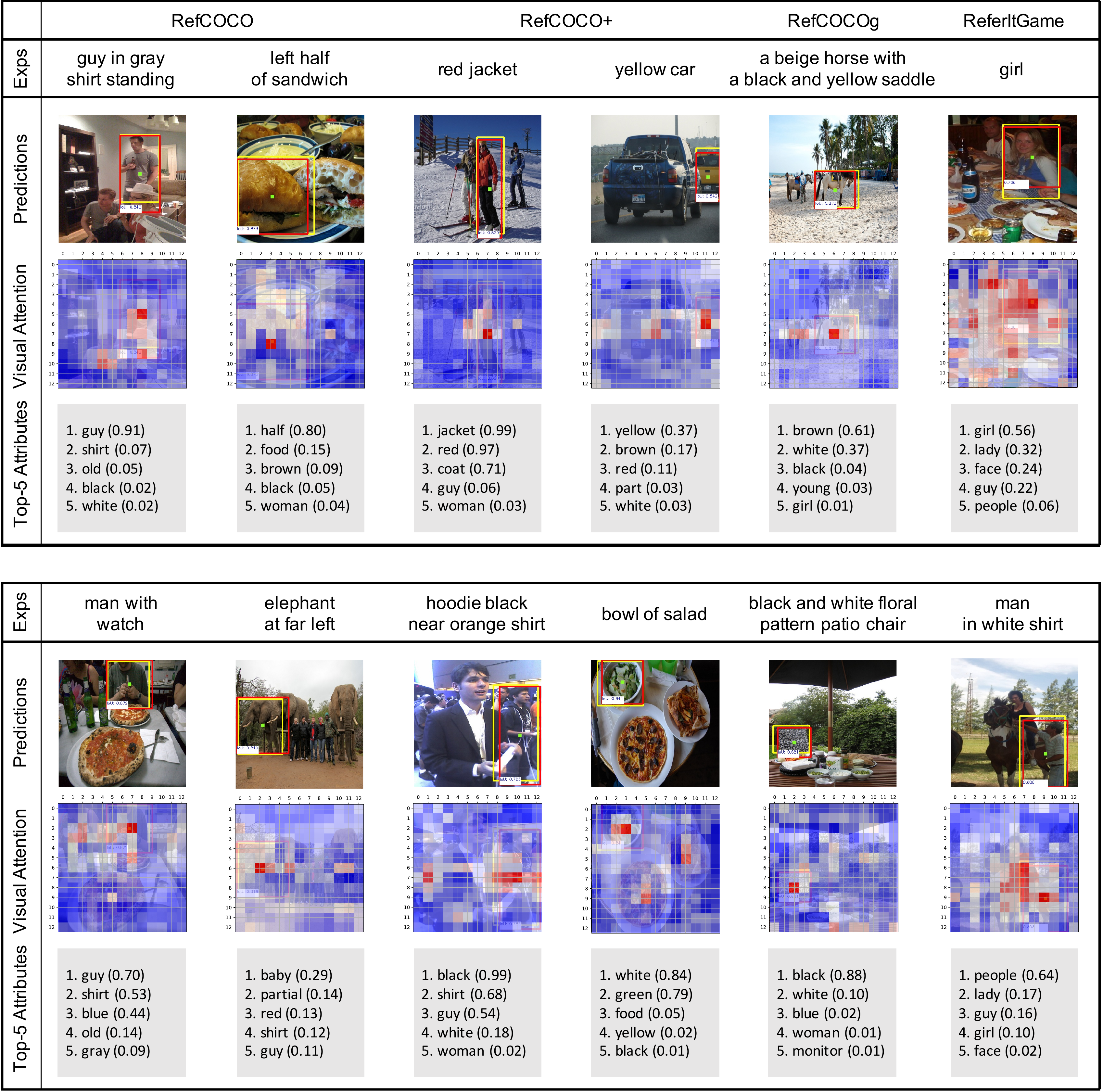}
  \caption{Qualitative results of the referring expression comprehensions with the corresponding visual attention heat maps and top-5 predicted attributes. The red rectangles denote the ground-truth bounding boxes, while the yellow ones denote the predicted boxes by our SSG. The green dots indicate the center points of the ground-truth bounding boxes.}
  \label{fig_qua4}
\end{figure}

\begin{figure}[ht]
  \centering
  \includegraphics[width=\linewidth]{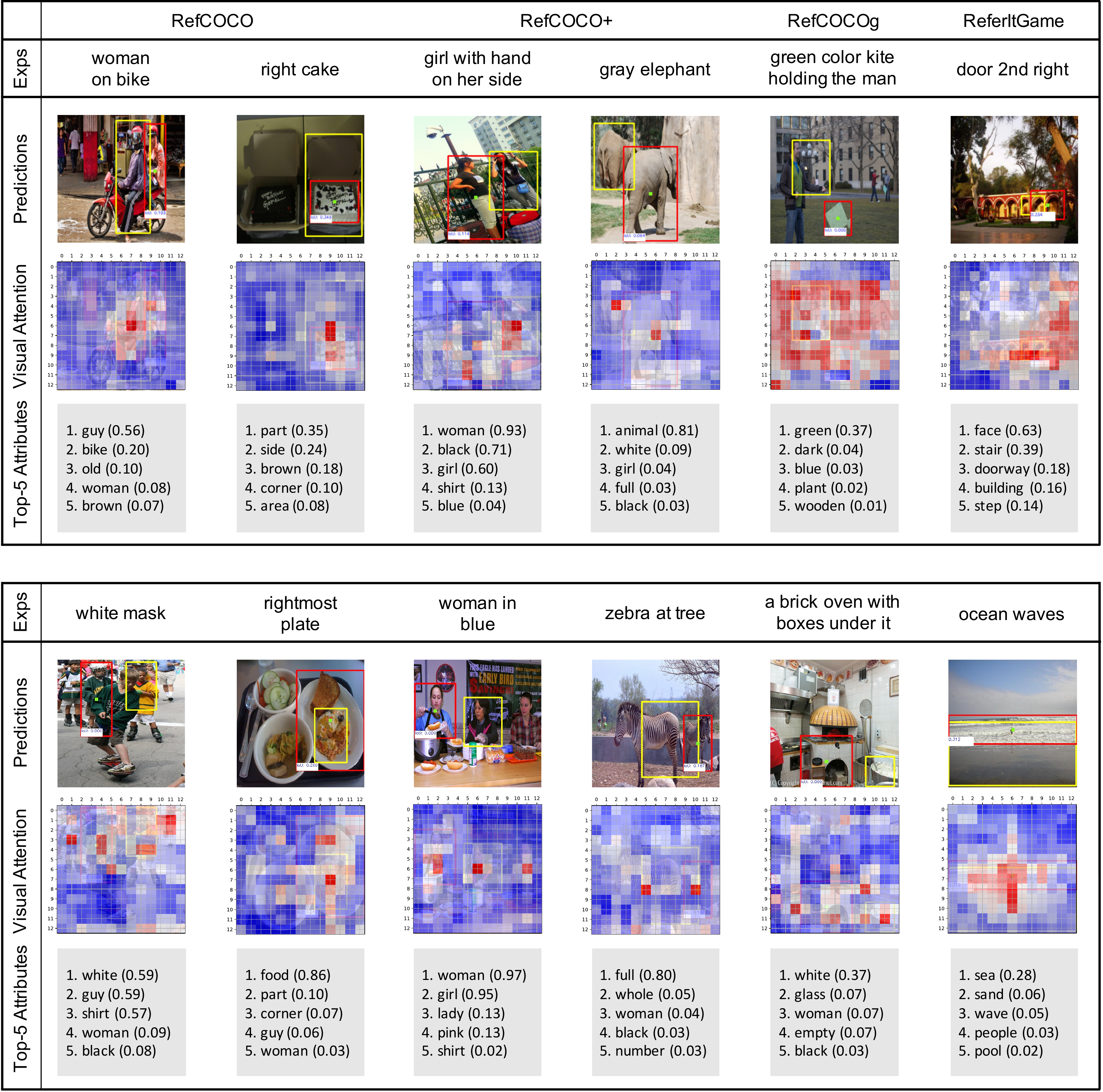}
  \caption{Some failure cases of the referring expression comprehensions with the corresponding visual attention heat maps and top-5 predicted attributes. The red rectangles denote the ground-truth bounding boxes, while the yellow ones denote the predicted boxes by our SSG. The green dots indicate the center points of the ground-truth bounding boxes.} 
  \vspace{-10pt}
  \label{fig_fail1}
\end{figure}

\begin{figure}[ht]
  \centering
  \includegraphics[width=\linewidth]{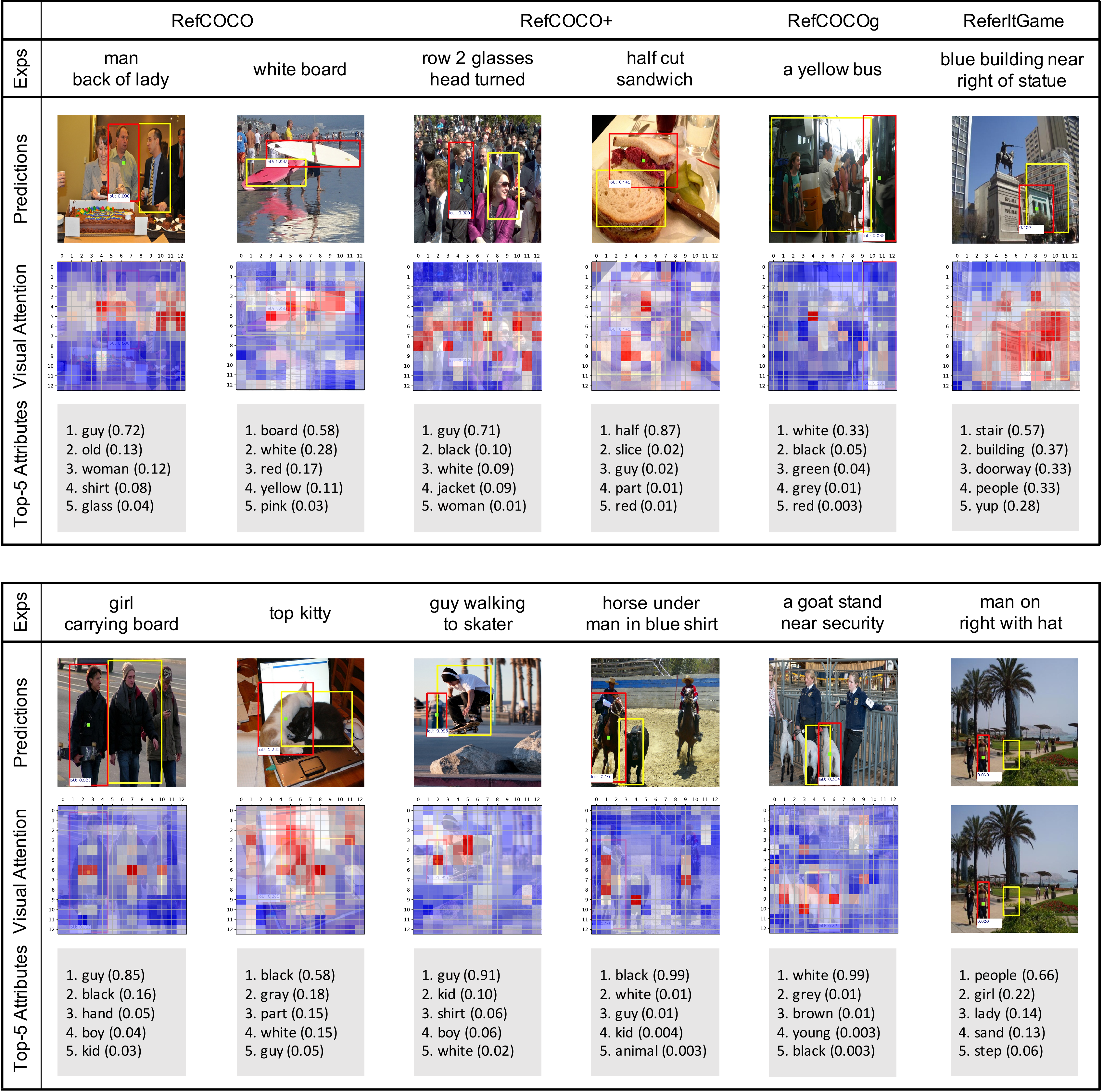}
  \caption{Some failure cases of the referring expression comprehensions with the corresponding visual attention heat maps and top-5 predicted attributes. The red rectangles denote the ground-truth bounding boxes, while the yellow ones denote the predicted boxes by our SSG. The green dots indicate the center points of the ground-truth bounding boxes.} 
  \vspace{-10pt}
  \label{fig_fail2}
\end{figure}

We show more qualitative examples of our SSG~($\lambda_{\text{loc} + \text{conf} + \text{att} + \text{attr}}$) in Fig.~\ref{fig_qua1}, Fig.~\ref{fig_qua2}, Fig.~\ref{fig_qua3}, and Fig.~\ref{fig_qua4}. As comparison, we also show some failure cases in Fig.~\ref{fig_fail1} and Fig.~\ref{fig_fail2}.

\subsubsection{Qualitative Results}

\clearpage
\subsubsection{Failure Cases}

\end{document}